\documentclass{article}

\usepackage[final]{ml4molecules_2022}

\usepackage[utf8]{inputenc} 
\usepackage[T1]{fontenc}    
\usepackage{hyperref}       
\usepackage{url}            
\usepackage{booktabs}       
\usepackage{amsfonts}       
\usepackage{nicefrac}       
\usepackage{microtype}      
\usepackage{xcolor}         
\usepackage{enumitem}
\usepackage{amsmath}

\usepackage{xspace}
\usepackage{amssymb}
\usepackage{graphicx}

\usepackage{array}

\newcommand{\researchQuestion}[1]{\textbf{(RQ#1)}\xspace}
\newcommand{\rqBenchmark}[0]{\researchQuestion{1}}
\newcommand{\rqCorrelation}[0]{\researchQuestion{2}}
\newcommand{\rqDistinguishability}[0]{\researchQuestion{4}}
\newcommand{\rqSimilarity}[0]{\researchQuestion{3}}
\newcommand{\rqErrorTypes}[0]{\researchQuestion{5}}

\title{Are VAEs Bad at Reconstructing Molecular Graphs?}

\author{%
  Hagen Muenkler\thanks{Correspondence to \texttt{hagen.muenkler\_ext@novartis.com} and \texttt{krmaziar@microsoft.com}}\\
  Novartis \\
  \And
  Hubert Misztela \\
  Novartis \\
  \And
  Michal Pikusa \\
  Novartis \\
  \AND
  Marwin Segler \\
  Microsoft Research AI4Science \\
  \And
  Nadine Schneider \\
  Novartis
  \And
  Krzysztof Maziarz \\
  Microsoft Research AI4Science \\
}

\begin{document}

\maketitle

\begin{abstract}
Many contemporary generative models of molecules are variational auto-encoders of molecular graphs. One term in their training loss pertains to reconstructing the input, yet reconstruction capabilities of state-of-the-art models have not yet been thoroughly compared on a large and chemically diverse dataset. In this work, we show that when several state-of-the-art generative models are evaluated under the same conditions, their reconstruction accuracy is surprisingly low, worse than what was previously reported on seemingly harder datasets. However, we show that improving reconstruction does not directly lead to better sampling or optimization performance. Failed reconstructions from the MoLeR model are usually similar to the inputs, assembling the same motifs in a different way, and possess similar chemical properties such as solubility. Finally, we show that the input molecule and its failed reconstruction are usually mapped by the different encoders to statistically distinguishable posterior distributions, hinting that posterior collapse may not fully explain why VAEs are bad at reconstructing molecular graphs.
\end{abstract}

\section{Introduction}

Generative modelling of drug-like molecules is an active area of research where an abundance of methods have already been developed~\citep{meyers2021novo}.
Machine Learning models, trained in an unsupervised way on a dataset of molecules, can successfully sample novel compounds that follow the same data distribution as the training set.
There is no consensus yet as to what representation suits the generation and optimization of drugs best, and thus while some methods use SMILES, a string-based representation of graphs, processing them with Recurrent Neural Networks (RNNs)~\citep{segler2017generating,winter2019learning,blaschke2020reinvent} or vanilla Transformers~\citep{fabian2020molecular,rothchild2021c5t5}, others use molecular graphs, employing Graph Neural Networks (GNNs)~\citep{jin2020hierarchical,maziarz2021learning} or Transformers adapted to graph data~\citep{maziarka2020molecule,maziarka2021relative,ying2021transformers,rampavsek2022recipe}.
Both representations have their strengths and weaknesses~\citep{zhu2021dual}. In particular, while models using RNNs on SMILES are typically more efficient, GNNs on molecular graphs give more fine-grained control over the generation process, allowing to incorporate valence or scaffold constraints~\citep{lim2019scaffold,li2019deepscaffold,imrie2020deep,maziarz2021learning,joshi20213d,yang2021hit}.

Among the graph-based approaches, one successful template is to use a Variational Auto-Encoder (VAE)~\citep{kingma2013auto}, built by pairing a GNN-based encoder with a sequential decoder which reconstructs the input graph step-by-step.
During training, the VAE trades-off reconstruction loss against a KL divergence term that ensures the encoded posterior does not deviate too much from the prior.
After training, one can produce new molecules by decoding samples from the prior, or find molecules with improved properties through latent space optimization~\citep{winter2019efficient}.

\looseness=-1
Although reconstruction is a central task in training VAE-based models, the downstream task of interest is usually sampling or molecular optimization, and thus reconstruction accuracy is often only briefly discussed.
Moreover, many papers use either a subset of ZINC~\citep{gomez2018automatic} or Polymers~\citep{st2019message} as their data source, and it is not clear how these reconstruction results generalize to a larger and more diverse dataset such as Guacamol~\citep{brown2019guacamol}.

In this work, we investigate three established VAE-based generative models of molecular graphs: JT-VAE~\citep{jin2018junction}, HierVAE~\citep{jin2020hierarchical} and MoLeR~\citep{maziarz2021learning}. We seek to answer the following research questions:

\begin{itemize}[nolistsep]
\item[\rqBenchmark] How well do these models perform on the reconstruction task if trained and evaluated using a large and diverse dataset?
\item[\rqCorrelation] Does reconstruction accuracy correlate with downstream capabilities, such as sample quality or optimization performance?
\item[\rqSimilarity] How similar are failed reconstructions to their ground-truth counterparts? Which aspects of the molecular graphs are preserved, and which are lost?
\item[\rqDistinguishability] Are latent embeddings of failed reconstructions too close to the embeddings of original inputs for the decoder to separate them?
\item[\rqErrorTypes] In which kinds of steps in the generation process are the models most likely to deviate from a valid reconstruction path?
\end{itemize}

We believe that answering these questions will shed some light on the capabilities of existing models, and help researchers interpret reconstruction accuracy results.

\section{Experiments}\label{sec:experiments}

\newcommand{\hiervae}[1]{HierVAE\textsuperscript{#1}}
\newcommand{\hiervaedrug}[0]{\hiervae{drug}}
\newcommand{\hiervaepoly}[0]{\hiervae{poly}}

We perform a series of experiments to answer each of the research questions outlined above. We compare JT-VAE, HierVAE, and MoLeR, using their open-source implementations and Guacamol~\citep{brown2019guacamol} as the source of training data. The HierVAE repository contains two model variants, one tuned for standard drug-like molecules and one for polymers; we evaluate both, referring to them as \hiervaedrug~and \hiervaepoly, respectively.

We compare the models under the following metrics: (a) reconstruction accuracy, measured on a sample of 10\,000 molecules from the test set; (b) Frechet ChemNet Distance (FCD)~\citep{preuer2018:fcd}, computed and transformed into the $[0,1]$ range by the Guacamol package; and (c) optimization performance when combined with Molecular Swarm Optimization (MSO)~\citep{winter2019efficient}, evaluated on the 20 optimization tasks from~\cite{brown2019guacamol} and reported as both raw score and compound quality measured using the filters from \cite{waltersQuality}. We defer further details to Appendix~\ref{app:metrics}.

\paragraph{\rqBenchmark Reconstruction accuracy across models and training duration}

\begin{figure}[t]
  \centering
  \includegraphics[height=4.24cm]{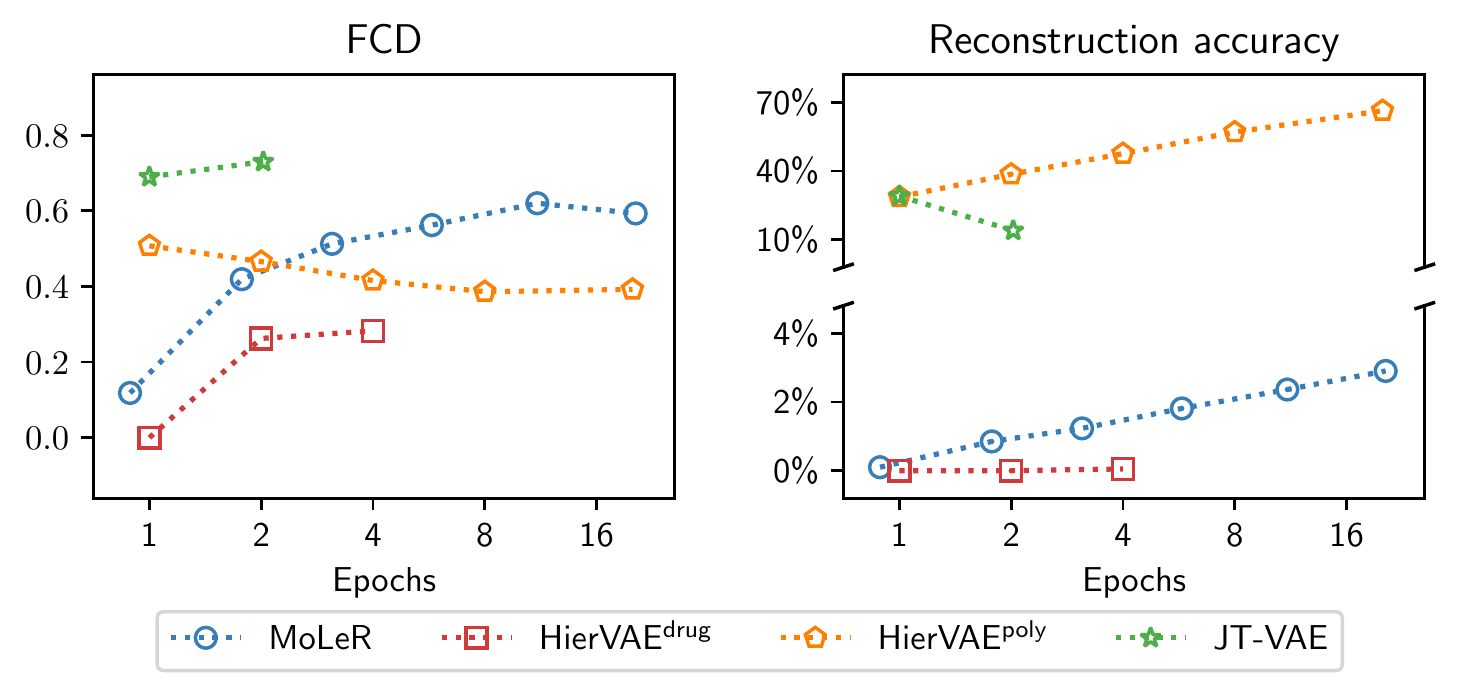}
  \includegraphics[height=4.24cm]{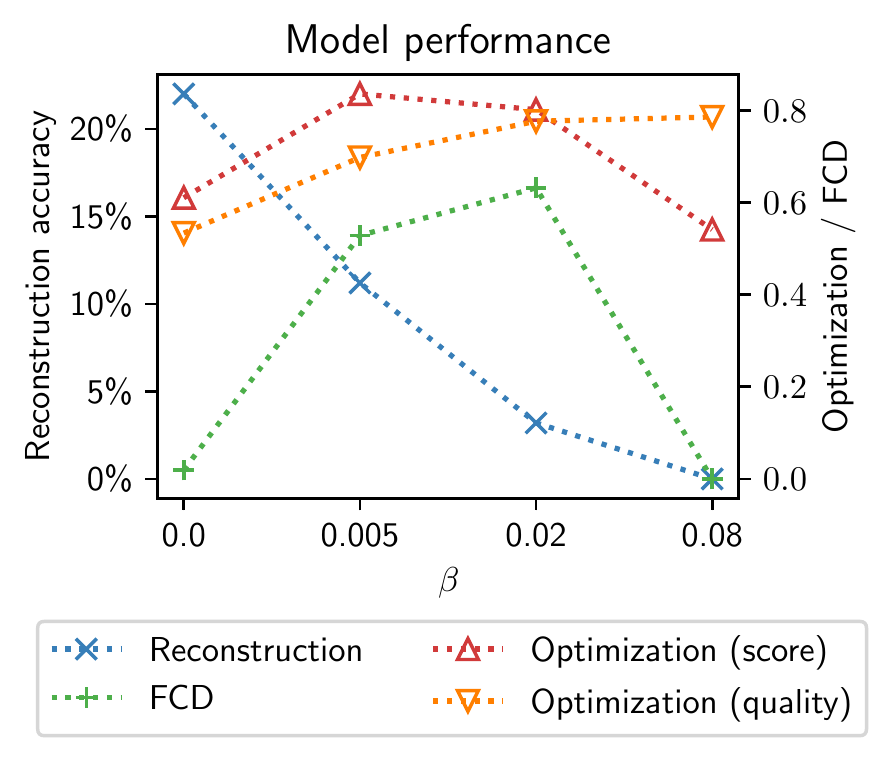}
  \caption{Left: FCD and reconstruction accuracy over the course of training for various model types. Right: Performance of MoLeR when trained with different values of the $\beta$ coefficient.}
  \label{fig:over-time-and-beta}
\end{figure}

For the first experiment, we train each model class and record the evolution of downstream metrics over time. As JT-VAE and \hiervaedrug~employ a schedule over the $\beta$ coefficient (controlling the loss contribution from the KL term), it is necessary to select the total number of training epochs up front; we experimented with various numbers of total epochs, scaling the schedules of other parameters proportionally. In contrast, MoLeR and \hiervaepoly~keep the $\beta$ coefficient constant (apart from a short ramp-up period in MoLeR), and so we train these models until convergence.

\looseness=-1
We found that the different model classes require a varying amount of time per epoch: 105 hours for JT-VAE, 27 hours for HierVAE, and 2 hours for MoLeR. For JT-VAE we capped the number of epochs to 2 due to resource constraints. For~\hiervaepoly \ we used the default training duration of $20$ epochs. For \hiervaedrug \ we tried various training lengths, but found training to be unstable as most runs suffered from posterior collapse, leading to decoder ignoring the encoder output; we report the sole successful run which trained for $4$ epochs\footnote{It is likely that re-tuning some hyperparameters or improving the schedule for the $\beta$ coefficient would make \hiervaedrug~train reliably on Guacamol.}. Finally, MoLeR was stopped due to convergence, with occurred after 20 epochs.

We visualize the results in the left part of Figure~\ref{fig:over-time-and-beta}. Training loss for all models goes down over time, but the trade-off between FCD and reconstruction behaves differently: while MoLeR improves on both metrics in tandem, \hiervaepoly~doubles its reconstruction accuracy in later epochs while sacrificing FCD, and JT-VAE achieves high reconstruction first (when the $\beta$ coefficient is small), and then gives up reconstruction to gain on FCD. It is also worth noting that the highest reconstruction accuracy (obtained by \hiervaepoly) is still only $66.28\%$, which is in stark contrast with models based on SMILES: for example, CDDD~\citep{winter2019learning} can successfully reconstruct around $95\%$ of the same molecules. \cite{jin2020hierarchical} measure reconstruction accuracy on polymers, and report $79.9\%$ for HierVAE and $58.5\%$ for JT-VAE - significantly higher than the results we observe on Guacamol, despite the polymer molecules being much larger. We hypothesize this could be due to larger chemical diversity in the drug-like space, which increases the number of valid ways in which motifs can be assembled.

\paragraph{\rqCorrelation Trade-off between reconstruction and other capabilities}

To verify whether reconstruction is a good proxy for model quality, we train several variants of MoLeR by varying the $\beta$ coefficient controlling the weight of the KL divergence term in the loss, and measure multiple downstream performance metrics. The results are shown in the right part of Figure~\ref{fig:over-time-and-beta}. With growing $\beta$, reconstruction falls monotonically, as the model is encouraged to encode less information in its posterior distribution; while reconstruction accuracy is $22\%$ for $\beta = 0$, it eventually drops to $0\%$ for larger $\beta$. In contrast, we find that metrics directly relevant in practice -- such as sample quality measured by FCD and optimization performance -- have a unique maximum for an intermediate value of $\beta$. Thus, reconstruction accuracy alone is not predictive of whether the model is useful for optimization. Finally, we note that quality of the optimized molecules increases monotonically with $\beta$, which is caused by decreasing coverage of chemical space: for large values of $\beta$, the model samples are of good quality but also highly repetitive. The default setting $\beta = 0.02$ used in \cite{maziarz2021learning} sacrifices reconstruction significantly, but is close to optimal on all other metrics.

\paragraph{\rqSimilarity Similarities between failed reconstructions and original inputs}

\begin{figure}[t]
  \centering
  \includegraphics[width=\linewidth]{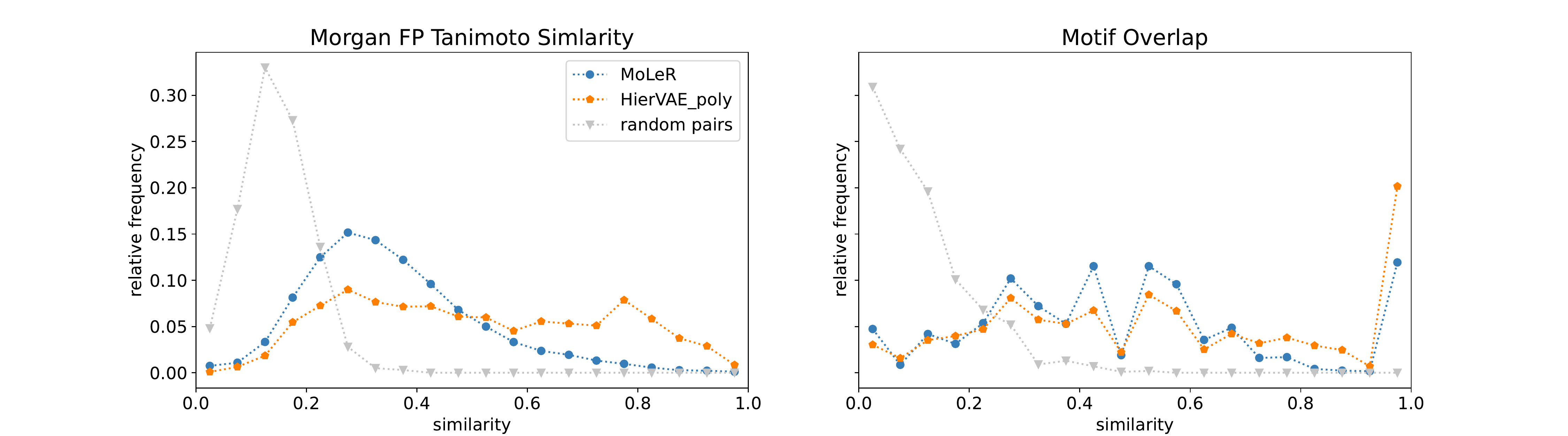}
  \caption{Similarities of reconstructions. In addition to the similarities of incorrect reconstructions to the original molecules for both \hiervaepoly~and MoLeR, we show a baseline of similarities for pairs of molecules drawn at random from our benchmark set.}
  \label{fig:similarities}
\end{figure}

In the case of a failed reconstruction, how similar are the molecules returned by the decoder to the molecules which were encoded 
originally? There are many notions of similarities between molecules,
here we consider the Tanimoto similarity between (count-based) Morgan fingerprints of radius 2~\citep{rogers:ecfp, rdkit} and motif overlap: each molecule is cut up into motifs using the algorithm described in~\cite{maziarz2021learning} to obtain a count-based motif-fingerprint. 

As seen in Figure~\ref{fig:similarities}, the failed 
reconstructions are significantly more similar than random pairs 
of molecules for both \hiervaepoly~and MoLeR. As expected from the higher 
reconstruction accuracy, the failed reconstructions for \hiervaepoly~are more similar to the original molecules than the ones generated
by MoLeR, with mean similarities of 48\% and 35\%, respectively. A similarity of 
35\% might not seem high, however, 99.7\% of random pairs are less than 35\% similar.
We observe a similar result for the motif overlap, where 
the MoLeR reconstructions have a mean similarity of 47\% 
compared to 55\% for HierVAE. It is interesting to note that while 12\% of the MoLeR reconstructions have exactly the same set of motifs as the original molecule, there are almost no cases with an overlap greater than 85\%, but not matching exactly. This is not the case for \hiervaepoly.
We defer further results to Appendix~\ref{app:extended-results-similarity}, where we also show that failed reconstructions are similar to original inputs in terms of solubility~\citep{schuffenhauser2020screening}.

\paragraph{\rqDistinguishability Distinguishability of latent space encodings}

The observed difficulty in reconstructing molecules might not only be due to problems in decoding, but 
may also be related to two molecules being encoded too close to each other in latent
space for the decoder to distinguish them.
In order to understand to what degree this contributes to the reconstruction accuracy, we compare the latent space encodings of the 
original molecules and their reconstructions. For a generic latent space, it is not
clear when two points are close. In the case of a variational auto-encoder, however,
the latent space is naturally equipped with a geometry, since we can measure how 
close two distributions are. Here, we propose a distinguishability measure that is based on the 
probability that an \emph{optimal} decoder can distinguish the two distributions (precise definition and more details are 
provided in Appendix~\ref{app:extended-results-distinguishability}). 

We note that random combinations of molecules almost always lead to distributions which are 
close to 100\% distinguishable. While the encodings of an original molecule and its reconstruction are 
closer, we observe that 55\% (89\%) are more than 97.5\% distinguishable for MoLeR (\hiervaepoly) (see Figure~\ref{fig:distinguish} in Appendix~\ref{app:extended-results-distinguishability}). While it is interesting to observe that \hiervaepoly~obtains a 
larger separation for false reconstructions than MoLeR does, 
we don't believe that the closeness of latent space encodings is the dominant factor limiting the reconstruction accuracy. 
We note however, that we are only comparing two latent space encodings here, while the model has to distinguish the given encoding from those of all molecules. Indeed, our analysis of varying the weight of the 
KL divergence shows that the organization of the latent space has a 
significant effect on the reconstruction accuracy.

\paragraph{\rqErrorTypes Types of errors in the stepwise generation process}

\begin{figure}[t]
  \centering
  \includegraphics[width=0.95\linewidth]{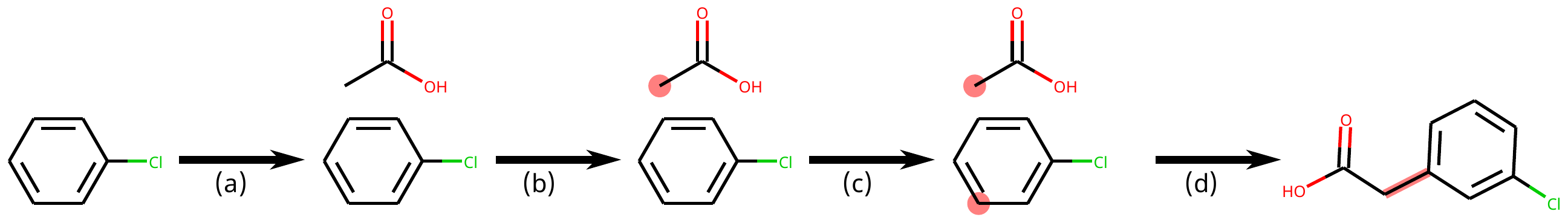}
  \caption{MoLeR's reconstruction steps for a sample molecule.}
  \label{fig:rec-steps}
\end{figure}

Noticing the low reconstruction accuracy, it is natural to ask why the generation 
fails so frequently. In order to approach this question, we study which kinds of 
errors MoLeR makes when reconstructing a molecule. This analysis is tailored to the specific reconstruction approach 
used in MoLeR and cannot immediately be transferred to other models, although a similar analysis could be carried out for these models as well. The specific reconstruction steps that MoLeR makes are 
summarized in Figure~\ref{fig:rec-steps}. Given a partially reconstructed molecule, MoLeR will first add an atom or motif (step a), then choose atoms to connect the fragments (steps b and c), and finally pick a bond type for the connection (step d). 

The analysis is based on finding the first step after which it is no longer possible 
for MoLeR to correctly reconstruct the molecule, and classifying the error that was made
in this step. 
We observe that more than 90\% of the errors are related to a failure in assembling the 
motifs, whereas it is rare that the model picks a motif that is not contained in 
the original molecule. 

It is interesting to contrast this finding with the results of the similarity analysis. 
In Figure~\ref{fig:similarities}, we observe that only 29\% of the reconstructions 
are composed of exactly the same motifs. We conclude that motifs which are not part of 
the original molecule are typically only added after an erroneous step has already 
occurred. The model is then decoding in a setup where the current partial graph 
is not a substructure of the original molecule; this scenario is not present during 
training. Indeed, the model will on average take 10 correct reconstruction steps before 
making an error. However, the model would need $49\pm25$ steps to correctly reconstruct
a molecule in our benchmark set, and hence a large part of the decoding takes place in
this situation. 

Details on the reconstruction steps and the error types 
we distinguish are provided in Appendix~\ref{app:extended-results-error-types}.

\section{Discussion}

We have analyzed the ability of JT-VAE, HierVAE and MoLeR to reconstruct molecules 
from their latent space encodings, comparing their reconstruction performance in the 
same train-test setup. 
The analysis of MoLeR's performance when varying the relative weight of the 
KL divergence term in the loss function showed that there is a trade-off between 
reconstruction accuracy and other performance measures.
We have further analysed which factors contribute to the lower reconstruction accuracy for MoLeR as compared to other models and concluded that the dominant 
factor are problems that the decoder has in recognizing how the different motifs 
are connected to each other. 
Inspecting the similarity between false reconstructions and the original molecules showed that these reconstructions still lead to related molecules with similar properties.

While our experiments show that increasing reconstruction accuracy naively by adjusting the loss weighing does not improve other capabilities of the model, we still believe that more fundamental modeling improvements targeting reconstruction will likely contribute to boosting the overall performance. Reconstruction provides the 
link between the actual molecules and their latent space encodings. Since all 
latent-space models depend on that, improving reconstruction should also improve 
the quality of these models, and lead to newly generated molecules that 
fit the desired property profile more closely. This is particularly 
important with respect to activity models, which are more sensitive to small modifications.

\begin{ack}

The authors would like to thank Paolo Tosco for very helpful discussions on the implementation of the error type analysis and Nikolaus Stiefl for insightful comments on the draft.

\end{ack}

\bibliography{references}
\bibliographystyle{abbrvnat}

\clearpage
\appendix

\section{Metrics}\label{app:metrics}

In this section, we explain in detail the metrics used in Figure~\ref{fig:over-time-and-beta}.

\paragraph{Reconstruction accuracy} We measure reconstruction by running a given molecule through the encoder and the decoder, and comparing the result to the original molecule by converting each to canonical SMILES. This process is deterministic, as we use greedy decoding and do not sample noise in the latent space (instead, we just pass through the mean of the encoded posterior). We measure reconstruction on a random sample of $10\ 000$ test set molecules, which we found to be enough to produce a precise estimate.

\paragraph{Frechet ChemNet Distance} FCD measures the extent to which samples from the prior of a trained model match the molecules from the training distribution. We use the implementation in the Guacamol package~\citep{brown2019guacamol}, which samples $10\ 000$ molecules from the prior, and compares them to $10\ 000$ sampled from the dataset. Both groups of molecules are passed through a pretrained ChemNet~\citep{preuer2018:fcd} feature extractor, and FCD is defined as the divergence between the network activations for the two sets of compounds. Guacamol then transforms this distance into the $[0,1]$ range, where higher FCD denotes following the data distribution more closely.

\section{Extended results}

\subsection{Similarities between failed reconstructions and original inputs \rqSimilarity}\label{app:extended-results-similarity}

We have discussed the similarity of the failed reconstructions based on a 
count-based Morgan fingerprint of radius 2 and a motif-based fingerprint. 
In both cases, the Tanimoto similarity is calculated according to

\begin{align}
    T_s (X, Y) = \frac{\sum  _{i} \min(X_i, Y_i) }
    {\sum  _{i} \max(X_i, Y_i)} , 
\end{align}
for two count-based fingerprints $X_i$ and $Y_i$. The computation 
of the motif-based fingerprint applies the algorithm outlined 
in \cite{maziarz2021learning}
to decompose a molecule into its motifs; it is not relevant whether 
the motifs obtained in this way are in the set of motifs used by MoLeR.

In addition to the different similarity measures, we also compare the 
properties of the reconstructed molecules to those of the originals. 
This is particularly relevant for latent space optimization (e.g. performed by running MSO in MoLeR's latent space, as in the experiments depicted in Figure~\ref{fig:over-time-and-beta}). 

\begin{figure}[h]
  \centering
  \includegraphics[width=0.75\linewidth]{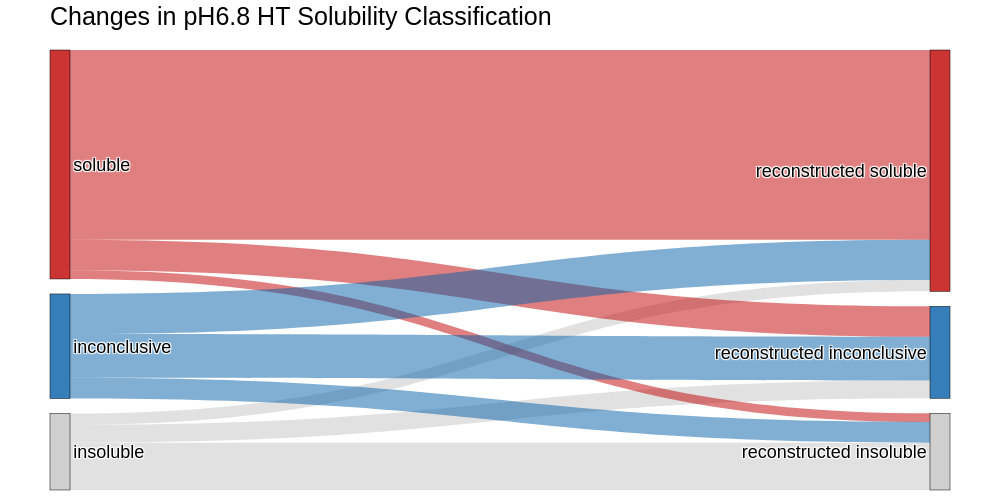}
  \caption{Comparison of solubility classification of reconstructed and original molecules.}
  \label{fig:solubility}
\end{figure}

Here, we consider the solubility of the molecules as predicted by 
the solubility model presented in \cite{schuffenhauser2020screening}, visualizing the results in Figure~\ref{fig:solubility}. Solubility is a key property in drug discovery for ensuring good bioavailability of a drug molecule. 
For the solubility classification, we observe that most reconstructions
are classified in the same way as the originals. Moreover, the changes that do occur are only rarely reversing the solubility classification. 
Typically either of the molecules is classified as inconclusive.

We note that while solubility might not vary too much between similar molecules, 
other properties, and in particular activity, can (see~\cite{stanley2021fs-mol} for an extensive discussion).
A detailed analysis of the property shifts between original and 
reconstructed molecules is left for future work.

\subsection{Distinguishability of latent space encodings \rqDistinguishability}\label{app:extended-results-distinguishability}

We expand our discussion on the question whether the latent space encodings of the molecule and its reconstruction are frequently 
too close to be distinguished by the decoder. 
In order to obtain an interpretable 
measure of the distance, we consider the following scenario: 
an \emph{optimal} decoder is tasked with deciding from which distribution a given point 
in latent space was drawn. Having full knowledge of those distributions, the decoder 
will choose the one that assigns a higher probability density to the latent space point. 
If we draw the point with equal probability from either of the two distributions, 
the probability that the optimal decoder will identify 
the correct distribution is given by
\begin{align}
    P_{opt} = \frac{1}{2} \int p(x) \theta \left( p(x) - q(x) \right) \mathrm{d}x +
    \frac{1}{2} \int q(x) \theta \left( q(x) - p(x) \right)\mathrm{d}x  ,
\end{align}
where $p$ and $q$ are two probability density functions, and $\theta$ denotes the 
Heaviside step function, which returns $1$ for positive arguments and $0$ for negative ones.

\begin{figure}[ht]
  \centering
  \includegraphics[width=0.65\linewidth]{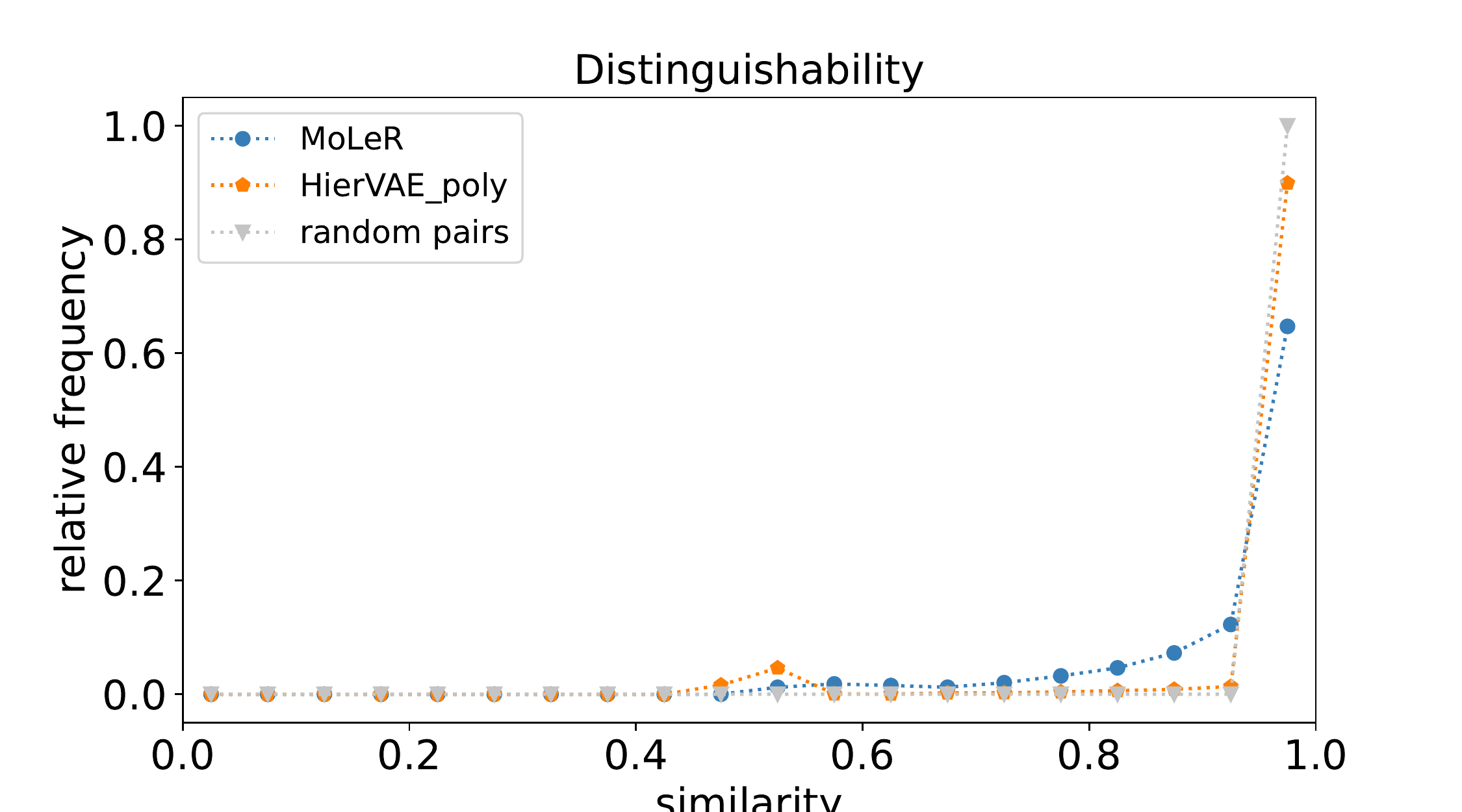}
  \caption{Distinguishability of encodings: We compare the distinguishability of the latent-space encodings of original molecules and their (false) reconstructions. In grey, we add the MoLeR encodings of random pairs of molecules. }
  \label{fig:distinguish}
\end{figure}

The results of applying this metric to the distributions obtained from the original molecules and their
reconstructions are shown in Figure~\ref{fig:distinguish}. We have included a baseline of the distinguishability 
of the MoLeR-generated latent-space encodings of random pairs of molecules taken from the benchmark set. We note that these can always be 
distinguished well by the optimal decoder, whereas the false reconstructions are closer in latent space, as expected. 
It is interesting to note that the false reconstructions obtained 
by \hiervaepoly~are more often easy to distinguish for the decoder 
than for MoLeR, despite the MoLeR latent space having more dimensions (512 compared to 24).
However, most of the pairs obtained by MoLeR's reconstructions can be distinguished with a high probability by an optimal decoder 
and we hence conclude that the closeness of latent space embeddings is likely not the dominant factor for the observed 
lower reconstruction accuracy.

\subsection{Types of errors in the stepwise generation process \rqErrorTypes}\label{app:extended-results-error-types}

In order to distinguish the different types of errors which occur in reconstruction, 
we recall the steps that MoLeR takes in reconstructing a molecule, visualized in Figure~\ref{fig:rec-steps}:

\begin{enumerate}[label=(\alph*)]
    \item Given a partial graph, the model first chooses an atom or motif to add.
    \item In the next step, the model selects an atom in the newly added motif to use to connect to the previous partial graph.  
    \item Then, the model selects an atom in the previous partial graph to connect to the newly added motif.  
    \item Lastly, the model chooses a bond type for the connection between the two previously selected atoms (one in the partial graph and one in the newly added motif). 
    \item At this point, the model can select to add another bond to the molecule. This step is repeated until the model decides to stop adding bonds.
\end{enumerate}

We note that steps (c) and (d) are actually implemented as a single step,
but for the purpose of disentangling types of reconstruction errors, we keep them separate here.
The last step allows the model to construct rings which are not present in the library of motifs 
that it uses. MoLeR builds kekulized molecules (which do not contain aromatic bonds, only sequences of single, double or triple bonds), so that an aromatic ring could be constructed as a 
sequence of single and double bonds. Correspondingly, in step (d), MoLeR chooses between single, 
double and triple bonds. 

The analysis of whether or not a step is correct is based on whether the partial graph at that 
step can still lead to a correct reconstruction. This is in turn mainly based on checking 
whether the current partial graph is a substructure of the original molecule. Since the partial 
graphs are kekulized, we also kekulize the original molecule and consider all resonance structures~\citep{rdkit} for the substructure search. 

The error types mainly correspond to the different steps outlined above.
In some cases, however, a step can fail in different ways.
For example, when adding a new motif, the step 
can fail either because the new motif is not contained in the original molecule, because it was already added in one of the previous steps, or because there is no way to attach it to the previously generated
graph to reach a valid substructure of the original molecule; cf. Table~\ref{tab:error_types} for 
more details on the error types and their frequencies for MoLeR. 

\begin{table}[ht]
    \centering
    \vskip0.4cm
    \begin{tabular}{
    >{\centering\arraybackslash} m{4cm} >{\centering\arraybackslash} m{1.6cm} >{\centering\arraybackslash} m{3.3cm} >{\centering\arraybackslash} m{3.3cm}}
         \textbf{Error Type} & \textbf{Frequency} & 
         \textbf{Original Molecule} & \textbf{Erroneous Step} \\ \hline
         Wrong attachment point & 52.4 \% & 
         \includegraphics[width=0.95\linewidth]{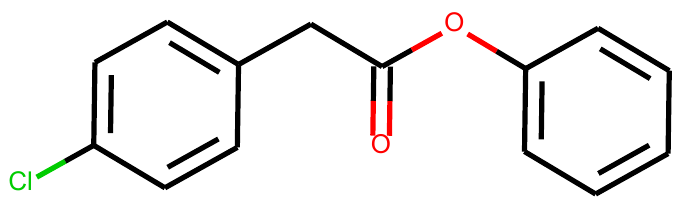} &
         \includegraphics[width=0.95\linewidth]{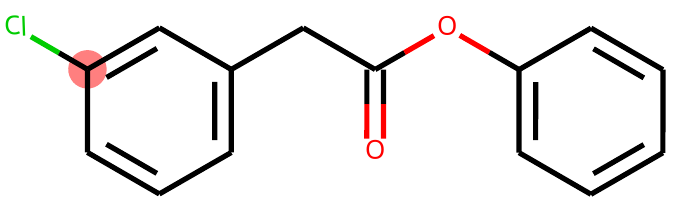} \\ \hline 
         New motif not attachable & 28.8 \% & 
         \includegraphics[width=0.95\linewidth]{figures/rec_example.pdf} &
         \includegraphics[width=0.95\linewidth]{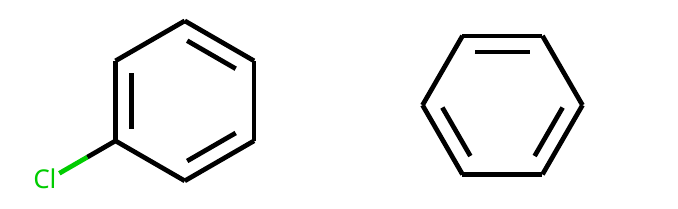} \\ \hline 
         Wrong bond type & 10.9 \% & 
         \includegraphics[width=0.95\linewidth]{figures/rec_example.pdf} &
         \includegraphics[width=0.95\linewidth]{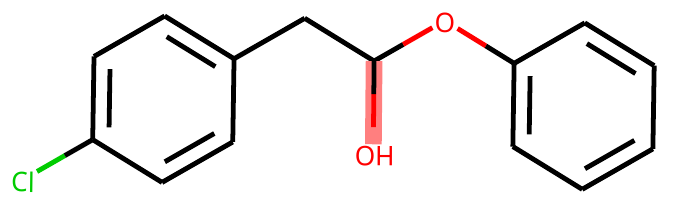} \\ \hline 
         New motif not contained & 4.5 \% & 
         \includegraphics[width=0.95\linewidth]{figures/rec_example.pdf} &
         \includegraphics[width=0.95\linewidth]{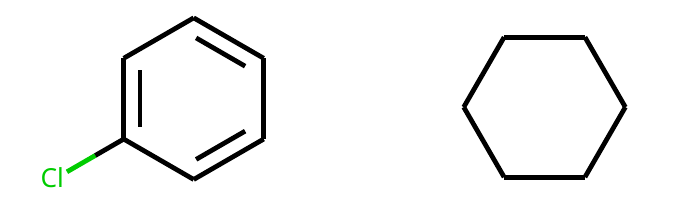} \\ \hline 
         Motif already added & 2.2 \% & 
         \includegraphics[width=0.95\linewidth]{figures/rec_example.pdf} &
         \includegraphics[width=0.95\linewidth]{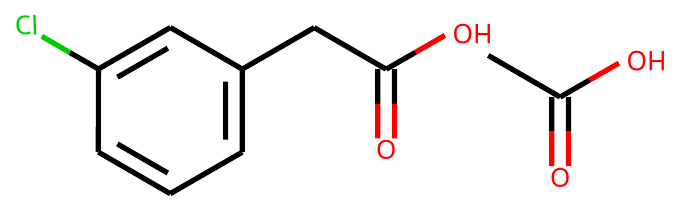} \\ \hline 
         Incorrect ring formed & 0.7 \% & 
         \includegraphics[width=0.95\linewidth]{figures/rec_example.pdf} &
         \includegraphics[width=0.95\linewidth]{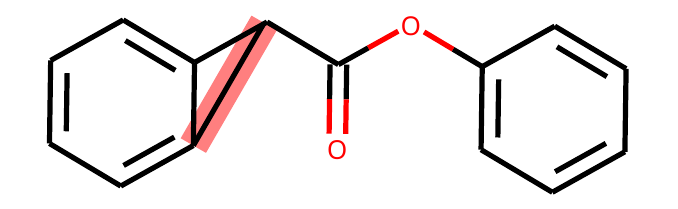} \\ \hline 
         First motif not in target & 0.5 \% & 
         \includegraphics[width=0.95\linewidth]{figures/rec_example.pdf} &
         \includegraphics[width=0.95\linewidth]{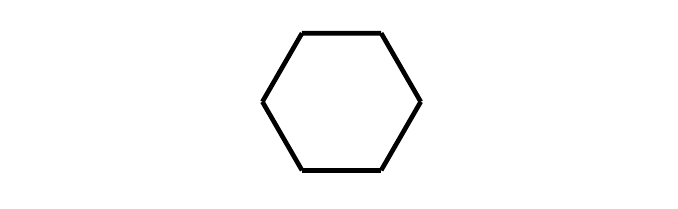} \\ \hline 
    \end{tabular}
    \vspace{5mm}
    \caption{Overview of error types and the frequency with which they occur for MoLeR reconstructions.}
    \label{tab:error_types}
\end{table}

The analysis shows that the main difficulty the model has in reconstruction is to understand the way in which the different motifs are connected. The two most frequent error types are both related to this problem, and together make up more than 80\% of the errors that the model makes.
The third most frequent error is also related to assembling the pieces correctly, and not to finding the collection of motifs which were present in the original molecule. 
Indeed, choosing a motif which was not part of the original molecule only occurs as the 
first error for 5\% of the reconstructions.

\end{document}